# Unsupervisedly Prompting AlphaFold2 for Few-Shot Learning of Accurate Folding Landscape and Protein Structure Prediction

Jun Zhang[1,†], Sirui Liu[1], Mengyun Chen[2], Haotian Chu[2], Min Wang[2], Zidong Wang[2], Jialiang Yu[2], Ningxi Ni[2], Fan Yu[2], Dechin Chen[3], Yi Isaac Yang[3], Boxin Xue[4], Lijiang Yang[4], Yuan Liu[5] and Yi Qin Gao[1,3,4,6,†]

**Affiliations:**

[1] Changping Laboratory, Beijing 102200, China.

[2] Huawei Hangzhou Research Institute, Huawei Technologies Co. Ltd., Hangzhou 310051, China.

[3] Institute of Systems and Physical Biology, Shenzhen Bay Laboratory, Shenzhen 518055, China.

[4] Beijing National Laboratory for Molecular Sciences, College of Chemistry and Molecular Engineering, Peking University, Beijing 100871, China.

[5] Department of Chemical Biology, College of Chemistry and Molecular Engineering, Peking University, Beijing 100871, China.

[6] Biomedical Pioneering Innovation Center, Peking University, Beijing 100871, China.

[†] To whom correspondence should be addressed. Email: jzhang@cpl.ac.cn or gaoyq@pku.edu.cn

**Abstract**

Data-driven predictive methods which can efficiently and accurately transform protein sequences into biologically active structures are highly valuable for scientific research and medical development. Determining accurate folding landscape using co-evolutionary information is fundamental to the success of modern protein structure prediction methods. As the state of the art, AlphaFold2 has dramatically raised the accuracy without performing explicit co-evolutionary analysis. Nevertheless, its performance still shows strong dependence on available sequence homologs. Based on the interrogation on the cause of such dependence, we presented EvoGen, a meta generative model, to remedy the underperformance of AlphaFold2 for poor MSA targets. By prompting the model with calibrated or virtually generated homologue sequences, EvoGen helps AlphaFold2 fold accurately in low-data regime and even achieve encouraging performance with single-sequence predictions. Being able to make accurate predictions with few-shot MSA not only generalizes AlphaFold2 better for orphan sequences, but also democratizes its use for high-throughput applications. Besides, EvoGen combined with AlphaFold2 yields a probabilistic structure generation method which could explore alternative conformations of protein sequences, and the task-aware differentiable algorithm for sequence generation will benefit other related tasks including protein design.

**Introduction**

According to the energy landscape theory[1-3], proteins live on a rugged landscape consisting of numerous local minima corresponding to various metastable conformations. Among all these local minima, there may exist a dominant one, and the corresponding conformation is often easier to be experimentally determined than others and attributed to biological activity. Strictly speaking, protein structure(s) is a probabilistic distribution conditioned on its sequence and the interacting contexts such as ions and solvents rather than a deterministic function of its sequence alone. Physics-based protein models, like force-field molecular dynamics[4], can in principle sample all function-relevant metastable conformations of a protein[5], but in practice this is almost impossible due to limited time budget. More direct and fast methods which can fold proteins into relatively stable structures are highly desired, giving rise to various protein structure prediction (PSP) algorithms.

In data-driven PSP, the structure label assigned to a sequence (usually determined by experiments) is often unique, thus, the mathematics is simplified to finding a function which can map a sequence of amino acids (denoted by $\mathbf{s}$) into a set of 3D coordinates (denoted by $\mathbf{x}$ throughout the paper). In most folding engines[6], this mapping is formulated as an implicit function $\mathbf{x} = \mathrm{argmin}_{\tilde{\mathbf{x}}} U_\theta(\tilde{\mathbf{x}}; \mathbf{s})$, where $U_\theta(\mathbf{x}; \mathbf{s})$ is a learnable energy (or score) function defining the folding landscape of the sequence. Such energy-based implicit modeling is also popular in other machine learning scenarios where iterative optimizations or multiscale complexities are involved[7-8]. The overall objective is usually dissected into two sub-tasks: i) defining a proper energy landscape, over which the global minimizer corresponds to the ground truth (i.e., target structure observed by experiments); ii) defining an efficient minimizer searcher (or "optimizer") which could find the target minimum across the energy landscape. The illustration of traditional PSP workflow is illustrated in Fig. 2a where an ideal convex folding landscape is assumed. However, due to the large dimensionality, we often need to deal with non-convex folding landscape as illustrated in Fig. 2b. Therefore, initial guesses and the choice of optimizer become crucial, and, for example, a template or reference structure may serve as a good initial guess and simplify the folding process[9]. On the other hand, constructing a smooth folding landscape based on available information is even more important for data-driven PSP. Many approaches have been developed to build the folding landscape according to physics priors or observed data. For example, direct coupling analysis (DCA)[10-11] over multiple sequence alignment (MSA) and associated approaches (e.g., GREMLIN[12-13]) aim to construct the folding landscape (or folding restraints) for the target sequence according to evolutionary homologs, whereas AlphaFold1[14] and related work[15-16] further showed that deep learning in combination with traditional co-evolutionary analysis can increase the accuracy and smoothness of the folding landscape. Historically, construction of folding landscape and finding the minimizer of the landscape were usually performed separately. With the advent of deep learning, efforts were made to fulfill these two tasks in an end-to-end manner such as energy-based models[17-18], RoseTTAFold[19] and AlphaFold2 (AF2)[20]. Nevertheless, how to solve the above-mentioned dual tasks still evidently influenced the design of end-to-end models. For example, the Evoformer module in AF2 mainly serves to learn the folding landscape based on evolutionary information[20-21], while the equivariant Structure Module plays a role of meta optimizer[22-23] given that recent studies have connected gradient descent with equivariant networks[24].

Different from most PSP models including AF1, AF2 takes raw MSA as input and does not require DCA or other statistics of MSA. In theory, a sufficient MSA depth which is essential for DCA is no longer necessary for AF2, and single-sequence PSP via AF2 is also possible. Although AlphaFold2 has raised up the baseline of PSP accuracy, however, it suffers significant drop of performance when MSA is limited[20]. Therefore, it evokes us to rethink the cause of such gap between the state-of-the-art model and the well-known Anfinsen's hypothesis[25] that the protein structure can be determined by its sequence. Besides, it is also appealing to investigate whether it is possible to close this gap, considering that once PSP can be made accurate with few or even without MSA, it would democratize PSP models without deploying resource-consuming and ever-growing sequence database, speed up the exploration of the protein universe, and help us better handle "orphan sequences". Besides, how to explore various conformations based on data-driven PSP models is still an open question[26], and it is to our interest that whether we can arrive at different local minima or metastable conformations by manipulating the folding landscape of end-to-end PSP models like AF2.

## Methods

### I. Deep probabilistic learning of MSA

The most common probabilistic model for MSA is arguably the Potts model[27-28], which describes the distribution of amino acids across MSA as a Markov random field, and is widely used in DCA and GREMLIN. Although Potts model belongs to the family of generative models, it has several limitations: It ignores any coupling between amino acids higher than the second order; Optimization of likelihood of Potts model involves calculation of the intractable partition function, so in practice specific gradient approximation methods like pseudo-likelihood are used; Worse still, a Potts model is only meaningful for a single set of MSA based on which the model is inferenced, and cannot directly transfer to another set of MSA. Recently, deep learning based approach was proposed to allow amortized optimization of Potts models across different MSAs[29]. However, it still assumes a pairwise coupling form and approximates the gradient using pseudo-likelihood. In order to optimize MSA, we developed a parametric probabilistic model for MSA, which overcomes the shortcomings of Potts model. Given a dataset (denoted by $\mathcal{D}$) containing many sets of MSA, each MSA set $m \in \mathcal{D}$ is defined for a center query sequence $Q_m$, and $m \equiv \{\mathbf{S}_m^i\}_{i=1,\ldots,N_m}$ contains $N_m$ aligned sequences with length of $L_m$ amino acids (note that $S_m^1 \equiv Q_m$). Similar to Potts model, our goal is to construct a statistical model $p_\theta$ ($\theta$ denotes optimizable model parameters) which maximizes the likelihood of the observed MSA,

$$\mathbb{E}_{\mathbf{S}\in m} p_\theta(\mathbf{S}) \equiv \mathbb{E}_{\mathbf{S}\in m} p(\mathbf{S}\mid\theta) \tag{1}$$

Note that unlike Potts model, the likelihood function Eq. (1) is written for the full-order joint distribution without any limited-order approximation. It is also different from masked language models like BERT[30] where only marginal distribution of masked amino acids in MSA is modeled[31]. There has been work leveraging Eq. (1) for a specific set of MSA[32]. However, since we hope our model can be transferable to different MSAs, we reformulate Eq. (1) using conditional probability in the form of Eq. (2),

$$\mathbb{E}_{m\in\mathcal{D}}\left[\mathbb{E}_{\mathbf{S}\in m} p_\theta(\mathbf{S}\mid Q_m)\right] \equiv \mathbb{E}_{m\in\mathcal{D},\mathbf{S}\in m} p_\theta(\mathbf{S}\mid Q_m) \tag{2}$$

To allow the model using more available conditional information, we further relaxed Eq. (2) into a multi-sequence conditional likelihood as conventionally used in meta generative learning[33-35],

$$\mathbb{E}_{m\in\mathcal{D},\mathbf{S}_m\in\{\mathbf{S}_m\}_{\text{target}}} p_\theta\left(\mathbf{S}_m\mid\{\mathbf{S}_m\}_{\text{context}}\right) \tag{3}$$

where we divide a full set of MSA $m$ into two (possibly overlapped) subsets: $\{\mathbf{S}_m\}_{\text{context}}$ serves as conditional information in Eq. (3), while $\{\mathbf{S}_m\}_{\text{target}}$ is used as training targets. Note that $Q_m \in \{\mathbf{S}_m\}_{\text{context}}$. Hence, it follows straightforwardly that Eq. (2) reduces to a special case of Eq. (3) when $\{\mathbf{S}_m\}_{\text{context}} = Q_m$. The key hypothesis underlying Eqs. (2-3) is the transferability or generalizability of MSA patterns. Researchers already know how to infer structural information from MSA patterns (i.e., $p(\mathbf{x}|\{\mathbf{S}\})$), and we in turn hypothesize that MSA patterns are subjected to a common implicit rule which can be learned by a transferable model. According to Bayesian theorem, $p(\{\mathbf{S}\}|\mathbf{x}) \propto p(\mathbf{x}|\{\mathbf{S}\})/p(\mathbf{x})$, this "hidden rule" is probably the 3D structure of proteins, and useful MSA patterns are indeed evidence of the protein structure during evolution. Following this reasoning, similar 3D structures may lead to similar MSA patterns, and our aim is to build a model which learns the relation from structure to MSA, and decodes meaningful structure-related MSA patterns.

## II. EvoGen: a hierarchical and differentiable generative model for MSA

Inspired by the success of autoregressive variational inference models[36-37], we derived a variational lower bound (Eq. (S3) in Supplementary Information) to tame the intractable likelihood in Eq. (3)[38]. The variational bound becomes tight when the expressivity of the probabilistic model is sufficiently large, so we adopt a deep neural network to model the conditional distribution in Eq. (3), and we named this deep neural network model as EvoGen. In the design of EvoGen, we sticked to two basic principles: "relativity" and "hierarchy". Given context sequences as input, the model is designated to learn the relative difference between targets and contexts in order to ease the training. Besides, it is well-known that sequences which are different at amino-acid level may be very similar in property or structure, therefore, it is reasonable to employ multiple feature spaces for amino-acid embedding in order to compare the relative difference between protein sequences. We thus introduced hierarchical feature spaces (or latent spaces) in EvoGen inspired by advanced deep generative models[37, 39]. Moreover, since we aim to model the full-length dependence between amino acids in a sequence, we do not use sequential autoregressive models which decode a sequence character by character in a given order. Instead, EvoGen is able to generate the whole amino acids of a sequence simultaneously through a diffusion-like generation process (Fig. 1a). EvoGen is composed of an encoder for inference and a decoder for generation. Both models are stacked by repeated Hyperformer blocks (Fig. 1c), which communicate between sequence and pair representations, and latent modules (Fig. 1d), which form statistics for context and target sequences. Overall, the inference and generation of EvoGen are fulfilled by a U-shaped[40] model which is widely adopted in modern diffusion probabilistic models[41]. More details about model architecture can be found in the Model Details in Supplementary Information.

## Results

### I. Shed light on the black-box folding landscape of AlphaFold2

In traditional PSP models, the depth, coverage, and diversity of MSA are known to influence the quality of DCA and the accuracy of the resulting folding restraints. However, since AF2 uses raw MSA as input without any explicit DCA-like feature extraction, little is known about how MSA influences the folding landscape of AF2, although we do know that the performance of AF2 drops dramatically as available MSA decreases[20]. Therefore, we designed an experiment to examine how AF2 responds to varied MSAs. We first selected a query sequence in CASP14 dataset[42], and randomly sub-sampled a certain number of MSA from the full MSA pool. We then fed these random MSA samples to AF2 and examined whether AF2 would produce varied structures for the same target. As Figure 2c shows, when we raised MSA depth up to 64 or more, AF2 is able to consistently produce a "converged" structure regardless of the randomness in the input MSA. However, intriguingly, the diversity of produced structures significantly increases as the number of MSA samples decreases. We further investigated this phenomenon using a larger test set containing 84 CASP14 targets (Fig. 2d; see Datasets in Supplementary Information for more details). It was confirmed that, with a sufficient large MSA depth, AF2 tends to fold the protein into converged structures regardless of the randomness in MSA samples. In contrast, given a small number of MSA, the folded structures produced by AF2 are particularly sensitive to the identity of the selected MSA. These findings are consistent with previous research, where AF2 has been deliberately implemented with fewer MSA in order to generate alternative conformations of G-protein-coupled receptors (GPCRs)[43].

This observation echoes the "maze hypothesis" proposed by Ovchinnikov et al.[44], where prediction of protein folding is analogized as finding the path throughout a maze, and homologue sequences share perturbed mazes with similar solutions, thus integrating mazes of a set of homologue sequences can lead to a "consensus maze" with smoother paths and easier solutions. Maze hypothesis can also be interpreted in terms of the well-known landscape theory for proteins[2-3]: Consider that one homologue sequence associated to a query has its own folding landscape consisting of multiple local minima, some of the local minima are unique to this individual sequence, but one or a few local minima (such as those corresponding to the native structure of the query) may be shared by most of the homologue sequences in MSA. Therefore, a possible mechanism of how AF2 constructs folding landscape based on MSA is that Evoformer manages to *integrate* the individual folding landscape of MSA sequences (Fig. 2e): Most of the local minima which are specific to few sequences are whitened or averaged out. On the other hand, those local minima shared by most of homologue sequences are eventually kept, resulting in a tractable folding landscape. But in practice we still do not know how to "optimize" MSA in order to make the folding landscape smoother, because MSA selection is not differentiable with respect to downstream goals. Besides, the "exploration-exploitation" dilemma was also observed in our experiments, where some converged structures produced at a high MSA depth are sub-optimal compared to certain structures occasionally produced with a smaller subset of MSA. This problem is reminiscent of what "prompt engineering" is trying to solve for state-of-the-art AI models in computer vision[45] and natural language processing[46]. It is thus desired to develop a data-driven MSA optimization strategy which can automatically prompt calibrated or useful MSA patterns to guide models like AF2 to fold better or explore alternative conformations.

**II. Unsupervised MSA calibration remolds the folding landscape**

In order to optimize MSA and smooth the folding landscape, we need a model to enhance the folding-relevant signals while whiten the disturbing ones in manually searched MSA. Therefore, we first curated a MSA dataset[47] which has good coverage and sufficient depth (see Datasets in Supplementary Information for more details). We then trained EvoGen on this dataset according to Eq. (3) or Eq. (S3) in an unsupervised manner (see Training Settings in Supplementary Information for more details). Provided that most of the MSA in the dataset contain folding-relevant signals, such an encoder-decoder scheme, as commonly adopted for denoising settings[48], can teach the model to reconstruct the common folding-relevant patterns while suppress the noisy or disturbing signals. During inference, EvoGen is able to transform a set of input MSA into less noisy output, and we term this transform as *MSA calibration*. Note that the number of MSA does not change during MSA calibration. Since the input and output of EvoGen are both MSA, it can be directly plugged into the inference workflow of AF2 without any fine-tuning or modifications. From this respect, Eq. (3) and EvoGen can be regarded as a new type of model-agnostic pretraining approach for PSP.

We then benchmarked EvoGen on a curated CASP14 test set (see Datasets in Supplementary Information for more details). To test whether our model can help smooth the folding landscape under poor MSA settings, we limited the number of MSA accessible by AF2 to be no more than 128 and ran all inference without templates. We made fair comparisons by running AF2 inference with and without MSA calibration when the same set of MSA were fed as input. Note that we turned off any settings which could cause non-deterministic effects during AF2 inference (see Inference Setting in Supplementary Information for more details). However, since EvoGen is a probabilistic model, it can yield varied output even if the input MSA is constant. We ranked the output structures according to the predicted confidence (i.e., averaged per-residue plDDT) and reported the most confident structure (called "first"). By convention, we also reported the best scored structure (called "best") assuming the ground-true score is known. Figure 3a shows that, given the same set of MSA, without any fine-tuning that requires structure labels, EvoGen could improve AF2 predictions over CASP14 targets.

Consistent with previously proposed mechanism (see Section I in Experiments & Results), we found that the improved performance is more significant for targets of small MSA depths. Therefore, it is appealing to check whether MSA calibration can help AF2 predict "hard targets" which naturally lack homologue sequences. Such targets are often termed as "orphan sequences", and we curated a "poor MSA" test set which consists of single protein chains with known PDB structures but with less than 30 available MSAs (see Datasets in Supplementary Information for more details). We then benchmarked EvoGen on this dataset following the same procedure as described for CASP14. From Fig. 3b it can be concluded that EvoGen is able to effectively improve AF2 predictions for targets of which the available MSA is noisy or insufficient.

We further compared the output structures of AF2 with and without MSA calibration. EvoGen can help AF2 predict the correct loop conformations using calibrated MSAs (Fig. 3c). Calibrated MSA can also help AF2 form correct secondary structures, for instance, from wrong helices to correct sheets as shown in Fig. 3d. In some cases (Fig. 3e), EvoGen even rescued AF2 from failed predictions by restoring the overall structures. We noticed that the restored parts of structure mostly correspond to sub-sequences with limited MSA coverage and regions of additional flexibility. This finding indicates that EvoGen may help smooth the folding landscape by promoting the folding-relevant signals which are noisy or less pronounced in the original MSA, thus reinforcing the target folding minimum.

Given a probabilistic model, we can now treat the "sequence-to-structure" problem from a probabilistic view (Eq. (4)) rather than the commonly used deterministic mapping:

$$\mathbf{x} \sim p(\mathbf{x}|\mathbf{s}) \Leftrightarrow \mathbf{x} = g(f(\mathbf{s};\mathbf{z})); \mathbf{z} \sim \mathcal{G}(\mathbf{0},\mathbf{I}) \tag{4}$$

where $f$ represents EvoGen, $\mathbf{z}$ is a random vector drawn from the standard normal distribution, and $g$

is a (deterministic) function such as AF2 which maps MSA to a 3D structure. Equation (4) enjoys a specific advantage that, by feeding different random Gaussian noises **z** to EvoGen, we can generate different structural conformations for a same sequence via AF2 even all the input sequences **s** (including MSA) are fixed. As introduced previously, during benchmark we generated several different structures (or decoys) for each target sequence by simply varying the random noises. For some target sequences, EvoGen plus AF2 led to different structural ensembles. Except for high-scored structures when compared to the ground truth, there may be some highly confident (according to pIDDT) but much lower-scored decoys. We hypothesized that such kind of highly confident decoys are likely to be alternative conformations of the target protein. Figure 4 provides the visualization of exemplary cases encountered during benchmark.

As an intriguing case, 3VNE (PDB code), which corresponds to protein VP24, is one of the eight proteins encoded by ebolaviruses[49]. VP24 is known to contribute to immune suppression and can bind host transcription factor STAT1[50]. EvoGen plus AF2 successfully generated an ensemble of predicted structures (Ensemble 1 in Fig. 4a) consistent with 3VNE which records the crystal structure of VP24 in an isolated monomer form. However, by virtue of probabilistic structure generation (Eq. (4)), we also observed another ensemble of predicted structures (Ensemble 2 in Fig 4a), which shows subtle but clear differences when superimposed with Ensemble 1. Particularly, there is a relative rotation and displacement of $\alpha$3-4 and $\beta$5-7 in Ensemble2 compared against the monomer structure (as indicated by the red arrows in the superposition of Fig. 4a). Intriguingly, according to the deuterium exchange experiment[50], $\beta$5-7 happens to reside near the hypothetic PPI interface between VP24 and STAT1. On the other hand, $\alpha$3-4 also shows dramatic changes in deuterium exchange rate after binding to STAT1[50], indicating the occurrence of local conformational changes. Therefore, structures in Ensemble2 which were not recorded in any PDB entry may possibly correspond to an alternative conformation of VP24 in the form of complex with other proteins like STAT1. In a similar case, given the same set of MSA but feeding varied random noises to EvoGen, we also obtained two remarkably different structure ensembles via AF2 for target 2X5T (PDB code). As shown in Fig. 4b, structures belonging to Ensemble 1 are highly similar to each other, and they also align well with the structure label in PDB, where two identical chains form an homodimer interfaced at the helix bundle[51] (highlighted by the red box in Fig. 4b). In contrast, another ensemble of highly confident structures (Ensemble 2 in Fig. 4b) was also observed. Structures in Ensemble 2 also align well with each other, and the main difference between these structures and those in Ensemble 1 lies in the overturn of the helix bundle (see the superposition in Fig. 4b) and the breakage of the PPI interface observed in the crystal structure. Considered that Ensemble 1 corresponds to conformations when the protein aggregate to dimers, Ensemble 2 may represent an alternative conformation when the protein takes an isolated or other complex form.

**III. Generative MSA augmentation stabilizes few-shot folding**

Given that calibrated MSA could improve the folding landscape, it naturally invokes the following question: Since searching and aligning natural sequences may cause disturbing noises in MSA, is it possible to directly "create" virtual MSA patterns which could form smooth folding landscape? This problem mirrors the well-known problem of inverse protein folding or protein design, where a sequence needs to be generated to stabilize or fit a specific protein structure. The subtle difference here is that we aim at generating MSA for a given sequence based on which a PSP model can easily fold the protein into the target structure. Similar to inverse protein folding, the generated MSA should also stabilize the target structure and create a relatively smooth folding landscape. Therefore, following the approach of actor-critic learning[52], we further trained EvoGen under the guide of AF2: EvoGen plays the role of actor or generator and randomly creates MSAs, whereas AF2 plays the role of critic to judge whether the generated MSA can help fold the target sequence into the correct structure. Thanks to the differentiability of virtual MSA generated by EvoGen, we can directly optimize parameters of EvoGen through back-propagation and the chain rule: We first computed the gradient of the loss functions of AF2 with respect to MSA, and multiplies the gradient of MSA with respect to the parameters of EvoGen. Therefore, EvoGen allows us to create MSA or protein sequences for the downstream objective in an end-to-end differentiable manner and optimize the model using any efficient first-order optimizers. To fine-tune EvoGen for PSP, we included supervised structural losses in AF2 (including FAPE and torsional angle losses) in our optimization objective. Besides, it is known that some sequences that are very different may mutually share a similar fold, and therefore, we regularized EvoGen to generate sequences not far away from naturally existing MSA patterns. This goal can be easily achieved by adding Eq. (3) into the final optimization objective, which encourages EvoGen to "imitate" how nature evolves protein sequences (see Training Settings in Supplementary Information for more details).

After fine-tuning EvoGen with the help of AF2, we tested whether EvoGen could generate MSAs which form reasonable folding landscape. We performed the benchmark following conventional procedures in few-shot learning: We gradually reduced the number of available MSA to AF2 and EvoGen, and compared the performance of AF2 with and without MSA generated by EvoGen. Indeed, EvoGen hereby is equivalent to an approach of data augmentation for AF2, so we term this procedure as *MSA augmentation* in order to mark its difference with MSA calibration. In MSA augmentation, EvoGen takes a small set of MSA as input but outputs a new and larger set of MSA.

We first benchmarked MSA augmentation on CASP14 test set. Consistent with the original paper of AF2, we found that by decreasing the number of input MSA, the performance of AF2 gradually drops (Fig. 5a). In contrast, with the augmentation of MSA provided by EvoGen, the overall prediction quality is relatively stabilized, and the drop of performance is effectively soothed. Particularly noteworthy, with only tens of MSA, EvoGen could keep the accuracy of AF2 near the same level as the full-MSA inference workflow. We also benchmarked the performance of EvoGen over the CAMEO test set (see Datasets in Supplementary Information for more details), and plotted the result in Fig. 5b, which led to the same conclusion as above. We further performed MSA augmentation for the poor-MSA test set and checked whether it can help improve the prediction of AF2 over "hard targets". Since sequences in this dataset have limited numbers of searched MSA, they can call trouble for AF2 inference. As shown in the scatter plot (Fig. 5c), given the same set of available searched MSA, many targets failed for original AF2 inference can now be accurately predicted with MSA augmentation. These experiments show that EvoGen could help AF2 achieve the state-of-the-art performance on naturally poor-MSA sequences.

We noticed that not all the targets we tested can be improved with MSA augmentation, so we analyzed and compared the structural characteristics of proteins where MSA augmentation can or cannot improve predictions. We first collected all the test sequences in the three datasets (CASP14, CAMEO and poor-MSA), and extracted two subsets: The "improved set" contains sequences for which MSA augmentation corrects the originally mis-folded predictions (AF2 TMScore<0.5 whereas EvoGen+AF2 TMScore>0.55); while the "underperformed set" consists of sequences for which MSA augmentation causes underperformance ($\Delta$TMScore<-0.05). The compositions of the secondary structures were calculated for these two sets (shown in cyan-colored columns in Fig. 5d). The overall structural compositions of the two sets are quite similar except that $\alpha$-helix is less redundant in the improved set (upper panel in Fig. 5d) compared to the underperformed set (lower panel in Fig. 5d). We also computed how much fraction of each secondary structure element was correctly predicted with and without MSA augmentation. For targets in improved set, all types of secondary structural elements were more accurately predicted with MSA augmentation. Particularly, some rare structures like $3_{10}$-helix and 5-helix were almost completely failed by

AF2 but were successfully restored with MSA augmentation. In contrast, we did not observe significant performance gap (except β-bridge) for underperformed targets in terms of secondary structures. This comparison indicates that MSA augmentation could generally improve the structural predictions of AF2 with limited risk of causing underperformance. We visualized several exemplary targets for which MSA augmentation significantly improved the prediction. In Figure 5e we compared the structures predicted by AF2 with or without MSA augmentation for three proteins with limited numbers of MSA. It can be seen that MSA augmentation can not only help improve structures of coil-abundant proteins like 4BFH (PDB code), but also correct the folding chirality of the anti-freeze protein 1Z2F (PDB code). Besides, the overall structure of a viral DNA polymerase 1T6L (PDB code) is also rescued by EvoGen.

**IV. How far are we from ideal single-sequence protein structure prediction?**

Finally it comes to a widely concerned question: Whether the structure of a protein can be accurately predicted merely by its sequence without using any other information like MSA or templates? Anfinsen's experiments showed that amino-acid sequence determines the stable structure of a protein. However, it is well-known that protein structures are only marginally stable and often dynamic[53], and many proteins have more than one metastable conformation, so the mapping from sequence to structure may not be deterministic. Consequently, single-sequence PSP can be rather complicated and challenging due to the high-dimensional nature of the folding landscape (see Fig. 1 for illustration), and most modern PSP models rely on additional information other than query sequence to reduce the complexity of such a high-dimensional non-convex optimization problem. Nevertheless, EvoGen provides a possible approach which formally enables MSA-based models to perform single-sequence PSP. Specifically, if no MSA information except for the query sequence is provided to EvoGen, the model can also generate MSA in a zero-shot manner, and the generated MSA can be fed into downstream PSP models such as AF2, leading to a non-deterministic single-sequence inference workflow.

We first conducted experiments using zero-shot MSA generation (or equivalently, single-sequence PSP) via EvoGen in combination with AF2 over CASP14 and CAMEO test sets. Figure 6a shows that, given merely the query sequence AF2 cannot predict the correct fold (defined as TMScore>0.5 by convention) for most of the targets in both test sets. However, if we provided AF2 with the MSA created by EvoGen through zero-shot generation, the overall quality of predictions was significantly improved, and more than a half of originally mis-folded structures were predicted with correct fold. This encouraging result promoted us to further test whether the zero-shot MSA generation plus AF2 works for natural "orphan" sequences. We then performed the same experiment over the poor-MSA test set and provided only the query sequence to EvoGen as conditional information. From Fig. 6b we can see that zero-shot MSA generation via EvoGen also significantly improved AF2 accuracy over these hard targets.

Since single-sequence PSP is not well-defined for proteins which exhibit multiple dynamically competing conformations[26], it is appealing to check whether EvoGen overfits the structure labels which often collapse multiple conformations into a specific one due to experimental conditions. We first divided the poor-MSA dataset according to the nature of the proteins into various categories, for example, whether the protein is natural or artificially designed, whether the protein is toxin or viral, and whether the protein structure is determined by X-ray or NMR. Intriguingly, we found that de novo proteins, viral proteins and toxins are specifically enriched in the poor-MSA dataset, indicating that these proteins naturally lack homologue sequences or co-evolutionary information. For each category of proteins, we computed how many AF2 predictions fall into correct fold (TMScore>0.5) with or without zero-shot MSA generation (Fig. 6c). We first observed that EvoGen can effectively help AF2 improve the quality of predicted structures for natural proteins. Particularly for viral proteins and toxins, the gains in performance are significant. Besides, EvoGen could help improve the quality of single-sequence PSP by a large margin for sequences whose structures are determined through X-Ray. In contrast, only limited improvement was observed for sequences whose structures are determined through NMR which often correspond to dynamic conformations. This finding suggests that EvoGen can help improve AF2 predictions for sequences which only have one dominant stable structure, but does not overfit for sequences which exhibit dynamic structures.

Intriguingly, we find that single-sequence AF2 is able to predict the correct fold for most of the artificially designed proteins even without the help of EvoGen, although EvoGen still helps raise the quality of predictions to a higher level. This finding can be reasonably explained by the fact that, compared to natural proteins, most of de novo proteins are hyper-stable because the sequence itself is deliberately designed to stabilize the structure. The folding landscape of such hyper-stable protein is usually much smoother and there only exists one dominant minimum across the energy landscape. Since MSA information for de novo designed proteins is usually limited due to their artificial nature, single-sequence structure prediction is often needed for protein designs. We thus further benchmarked zero-shot EvoGen plus AF2 on a more commonly used de novo protein dataset[54-55]. As Fig. 6d shows, although the original AF2 already outperforms existing methods like RaptorX[55] in single-sequence PSP setting, AF2 still fails for some targets and yields wrong folds (TMScore<0.5). With the help of zero-shot MSA generation, all the targets can be predicted into correct folds and the overall performance reaches state of the art. In Fig. 6e, we visualized two de novo targets where the original single-sequence AF2 predictions failed. 6CZG is an artificially designed β-barrel protein, but without zero-shot MSA augmentation, AF2 cannot predict its correct topology and chirality. As the second case, 6W3F is an artificially designed enzyme-like protein containing a binding pocket, but AF2 wrongly predicted its structure, especially that of the key pocket, while EvoGen helps restore the correct fold and accurately predict the pocket structure.

**Discussion**

Fast and accurate PSP models are of great practical value because they make efficient exploration of protein sequence space possible, particularly in design applications[56-57]. The advent of AlphaFold2 has raised the bar for the accuracy of data-driven PSP, and it has been assisting researchers to expand the database of proteins since its birth[58]. AF2 is different from most of previous PSP models in that it is end-to-end, and that it inputs raw MSA features without any co-evolutionary analysis. In theory, AF2 is able to perform few-shot-MSA or even zero-shot-MSA (i.e., single-sequence) PSP. Without deploying resource-consuming and ever-growing sequence database, single-sequence PSP is quite appealing because it could democratize PSP models for large-scale and high-throughput applications, and dramatically speed up the exploration of the protein universe. Besides, currently about 1/5 of all metagenomic protein sequences[59] and about 11% of eukaryotic and viral proteins[60] are estimated to be "orphan" which naturally lack sequence homologs. Dealing with these sequences requires PSP models to make accurate predictions with very limited evolutionary information. Unfortunately, the performance of AF2 is guaranteed only if the available MSA is sufficiently deep. Unlike DCA where large MSA depth is known as prerequisite, how MSA influences AF2 and why shallow MSA harms its performance is quite obscure. In this paper, we tried to open this "black box", investigated the mechanism of how AF2 constructs folding landscape from provided evolutionary information, and proposed a simple but tractable physical picture explaining the observed connections between MSA and AF2 performance.

Assuming that MSA is evidence of 3D structures, we designed a deep probabilistic model for MSAs called EvoGen. Different from Potts model, EvoGen is a full-order joint probabilistic model as well as a meta-generative model for MSA. It is designed to learn generalizable features across MSAs for different query sequences. We designed a specific U-shaped neural network architecture for EvoGen so that it is transferable to MSAs with varied depth and length. We also formulated a variational lower bound for the full-order joint likelihood so that the training and inference of EvoGen can be performed efficiently.

Inspired by the success of prompt engineering in modern deep learning[45-46], two plug-in methods were developed on the basis of EvoGen, which can prompt MSA-dependent models with calibrated or virtually generated homologue sequences, hence, improving the accuracy of PSP models like AF2 when dealing with poor MSA targets. On the one hand, serving as an unsupervised data-denoising strategy, MSA calibration is able to effectively whiten or denoise the manually searched MSA, thus correct the folding landscape and help AF2 fold better. To achieve MSA calibration, EvoGen is trained only on the sequence database without any structure labels, and the downstream AF2 is not fine-tuned at all. On the other hand, MSA augmentation provides virtually generated MSA to downstream AF2 as data augmentation which can stabilize the prediction quality in low-MSA regime. Particularly, zero-shot MSA augmentation can help AF2 improve single-sequence predictions over hyper-stable proteins. To achieve MSA augmentation, we trained EvoGen under the guide of AF2 over a limited number of structure labels, and we can directly backpropagate the structural losses of AF2 to EvoGen owing to the differentiability of the generated MSA. Moreover, by functional compositions of the probabilistic EvoGen and a deterministic function like AF2, we obtained a new type of probabilistic end-to-end PSP algorithm, which could yield varied structures given a unique input. We showed that such probabilistic PSP algorithm could help explore alternative conformations of proteins which can be crucial for drug discovery.

Generally speaking, EvoGen can be regarded as a model-agnostic unsupervised pre-training strategy for protein-related tasks. Different from other language modeling (LM) pre-training like BERT, we showed that EvoGen can even work well with limited change of downstream models. This merit arises from the fact that the output of EvoGen is MSA, hence can be directly incorporated by MSA-dependent downstream models. This is particularly beneficial in that unlike natural language processing, protein-related model is usually very large and complicated (e.g., AF2), so the fine-tuning is usually more difficult than pre-training itself. On the other hand, it is also possible to employ MSA generation instead of masked language modeling as the pre-training task and feed the latent representation learn to the downstream PSP model. As recent study shows, reasonable single sequence predictions can be made when a PSP model is trained entirely on single-sequence data while taking BERT as an auxiliary or pretraining task[61-62]. It is thus appealing to investigate whether training a PSP jointly with zero-shot EvoGen instead of BERT can achieve better single-sequence predictions and we leave this study to future. Besides, as a differentiable sequence generator, EvoGen can be trained straightforwardly according to downstream objectives, and generate optimized protein sequences efficiently without performing Monte Carlo or gradient descent. Therefore, we expect the model and algorithm behind EvoGen can also assist other sequence generation tasks like protein design[57] as well as MSA-based protein learning tasks such as functional annotation and mutation assessment[63] *etc.* in the future.

**Acknowledgments**

This work was supported by National Key R&D Program of China (No. 2022ZD0115001), National Natural Science Foundation of China (22050003, 92053202, 21821004, 21927901 to Y.Q.G.) and CAAI-Huawei MindSpore Open Fund to J.Z. The authors thank Dr. Xing Che, Dr. Piya Patra and Yuhao Xie for useful discussion, and gratefully acknowledge the support of Xi'an Future Artificial Intelligence Computing Center, MindSpore, CANN (Compute Architecture for Neural Networks) and Ascend AI Processor used for this research.

**Supplementary Information**

The derivation of the variational lower bound for the log-likelihood adopted by EvoGen can be found in Supplementary Information. Supplementary Information also contains the detailed description of the training and test datasets present in this work and the model architecture as well as the settings for training and inference of EvoGen.

**Code & Data Availability**

Codes of EvoGen as well as associated model checkpoints have been released at MindSPONGE platform (https://gitee.com/mindspore/mindscience/tree/master/MindSPONGE/applications/MEGAProtein). Test datasets and associated benchmark results are also released.

**Conflict of Interest**

Huawei Technologies Co. Ltd. And Changping Laboratory are in the process of applying for provisional patents (Huawei Technologies Co. Ltd.) covering generative MSA learning for protein structure prediction, that lists J.Z., S.L., M.C., H.C., M.W., Z.W., J.Y., N.N., F.Y. and Y.Q.G. as inventors. The other authors declare no conflict of interest.

**Figures & Legends**

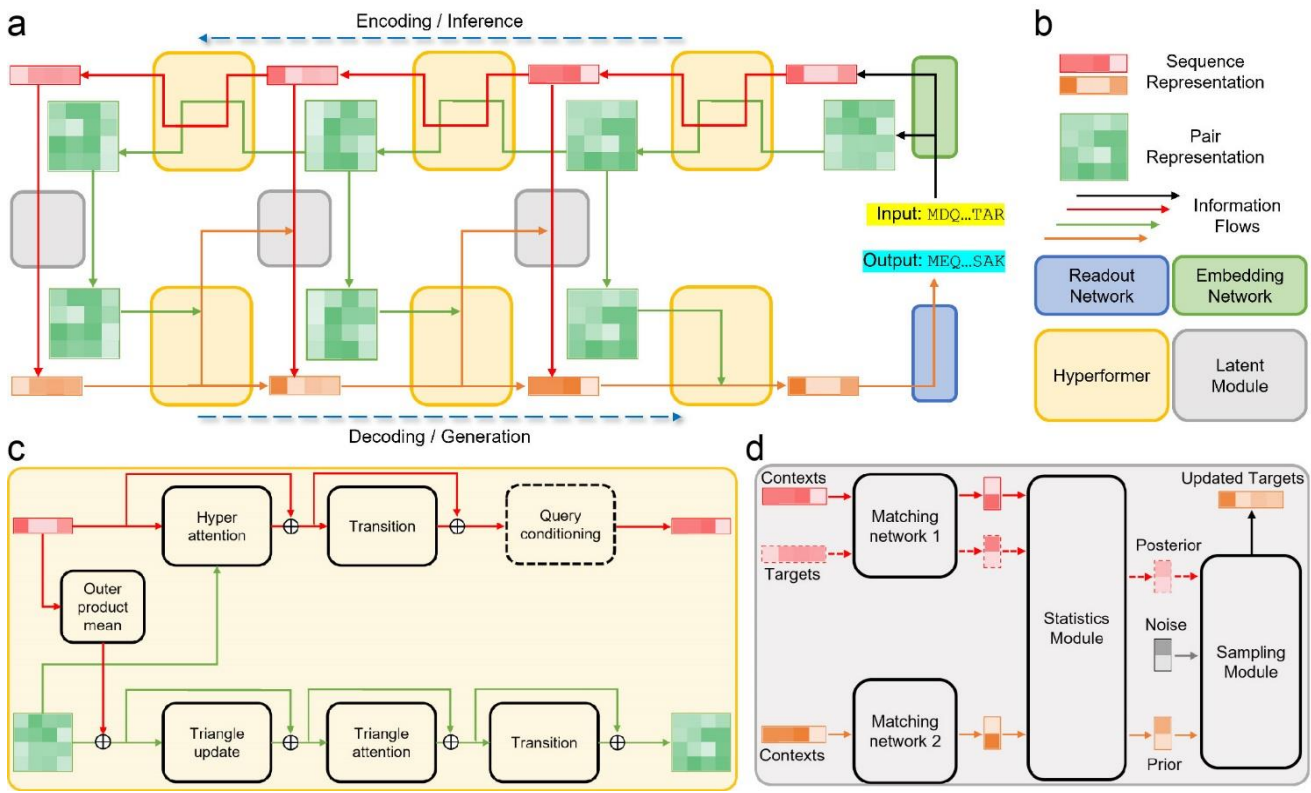

**Fig. 1 | Illustration of EvoGen model architecture. a**, Overview of UNet-like architecture of EvoGen. The input and output of EvoGen are both amino-acid sequences. **b**, Main components and building blocks of EvoGen. EvoGen simultaneously learns sequence and pair representations, and they interact with each other through Hyperformer. Latent Module performs statistics over MSA. Embedding Network transforms sequences into vector space while Readout Network transform vector representations back to sequences. **c**, The illustration of detailed inner logics of Hyperformer. **d**, The illustration of detailed inner logics of Latent Module. All parts shown in dashed boxes in **c** and **d** are only present in Encoder during inference but not in Decoder during generation.

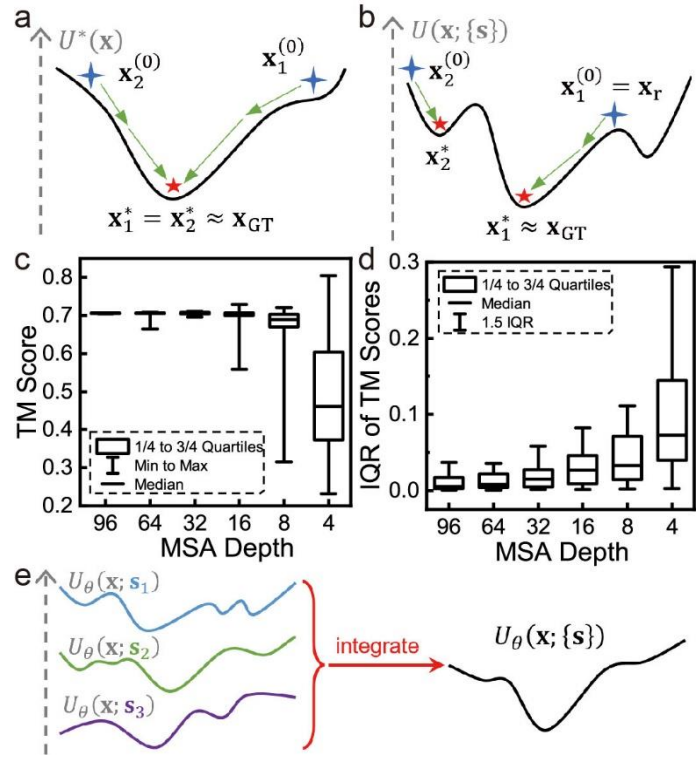

**Fig. 2 | Characterizing folding landscape of AF2. a**, Illustration of an ideal convex folding landscape $U^*(\mathbf{x})$ for PSP. Due to the convexity, different initial guesses, $\mathbf{x}_1^{(0)}$ and $\mathbf{x}_2^{(0)}$, all lead to the same optimal structure $\mathbf{x}^*$ which falls near the ground truth $\mathbf{x}_{GT}$. **b**, Illustration of a real-world folding landscape in PSP, where multiple local minima coexist, and different initial guesses $\mathbf{x}_1^{(0)}$ and $\mathbf{x}_2^{(0)}$ may lead to different (sub)optimal structures. Reasonalbe initial guess according to some reference or template structure $\mathbf{x}_{ref}$ could help find the correct minimum. **c**, TMScores of AF2 predictions for T1032 in CASP14 with randomly sub-sampled MSA at varied MSA depth. 48 independent trials were performed for each MSA depth, and the result was reported in box plot. **d**, Distribution of the inter-quartile range (IQR) between 1/4 quartile (Q1) and 3/4 quartile (Q3) for all sequences in CASP14 test set at varied MSA depth. For each sequence we performed the same random trials as in panel **c** and collected its IQR of TMScores. The IQRs of all applicable targets at a given MSA depth were compiled in the form of box plot. Unless stated otherwise, the hat lines of box plots correspond to (Q1-1.5IQR) and (Q3+1.5IQR) respectively. **e**, Illustration of the proposed mechanism of how AF2 extracts folding landscape according to evolutionary information. Each sequence homolog in MSA has its own sequence-dependent folding landscape $U_\theta(\mathbf{x}; \mathbf{s})$, and AF2 manages to integrate these individual landscapes into a new one $U_\theta(\mathbf{x}; \{\mathbf{s}\})$, where most of the noisy and disturbing local minima are averaged out and keeps only the target minimum commonly shared by all homologue sequences.

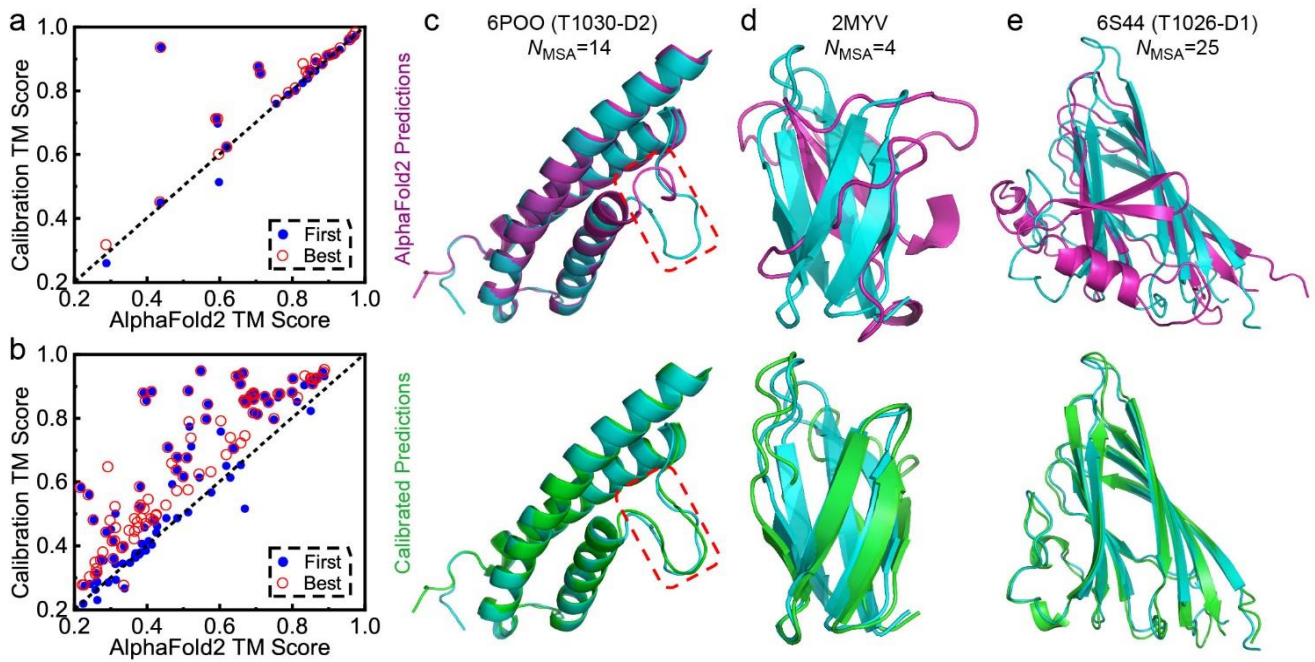

**Fig. 3 | MSA calibration via EvoGen improves predicted structure quality. a**, The quality of predicted structures with and without MSA calibration over CASP14 test set. Multiple random trials of MSA calibration were performed for a given set of MSA. Solid blue circles represent the top-ranked predictions according to confidence; Hollow red circles represent the best scored predictions given ground truth. Only data points with |ΔTMScore|>0.005 are shown for better visualization. **b**, The quality of predicted structures with and without MSA calibration over CAMEO test set. Only data points with |ΔTMScore|>0.05 are shown for better visualization. Data points are colored the same as panel a. **c~e**, Comparison of structures predicted for different targets (6POO or CASP14 T1030-D2 in **c**, 2MYV in **d**, 6S44 or CASP14 T1026-D1 in **e**) without MSA calibration (magenta), with MSA calibration (green), and the ground truth (cyan). Red dashed box serves to guide the view.

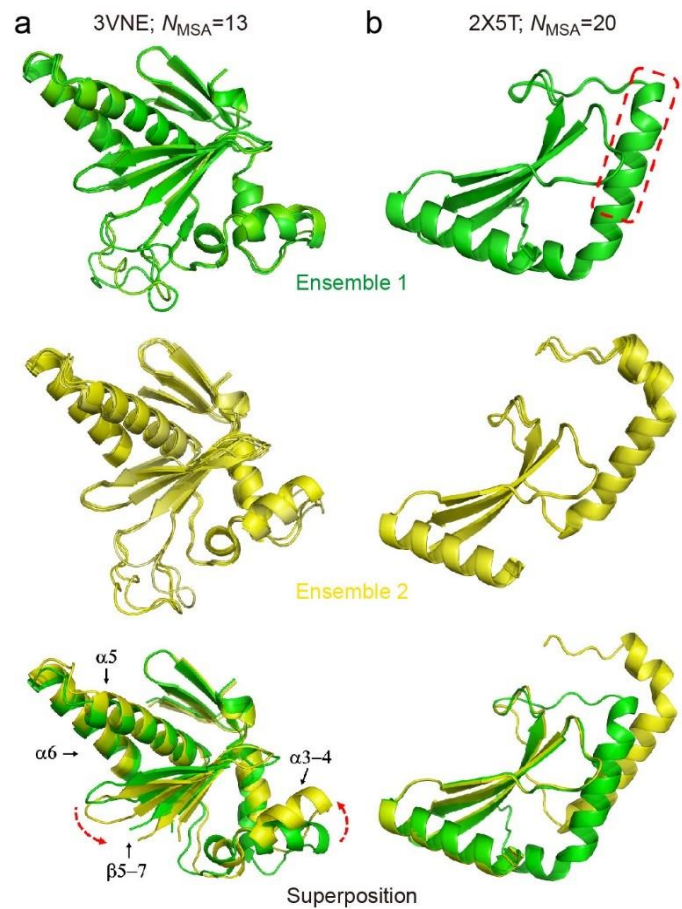

**Fig. 4 | EvoGen helps AF2 explore different protein conformations.** Predicted structural ensembles for 3VNE (PDB code) in **a** and 2X5T (PDB code) in **b**. Three randomly predicted structures were aligned for each ensemble. Ensemble 1 (upper panel) is close to the ground truth; Ensemble 2 (middle panel) correspond to alternative conformation. The superposition (lower panel) compares the differences of the two conformations.

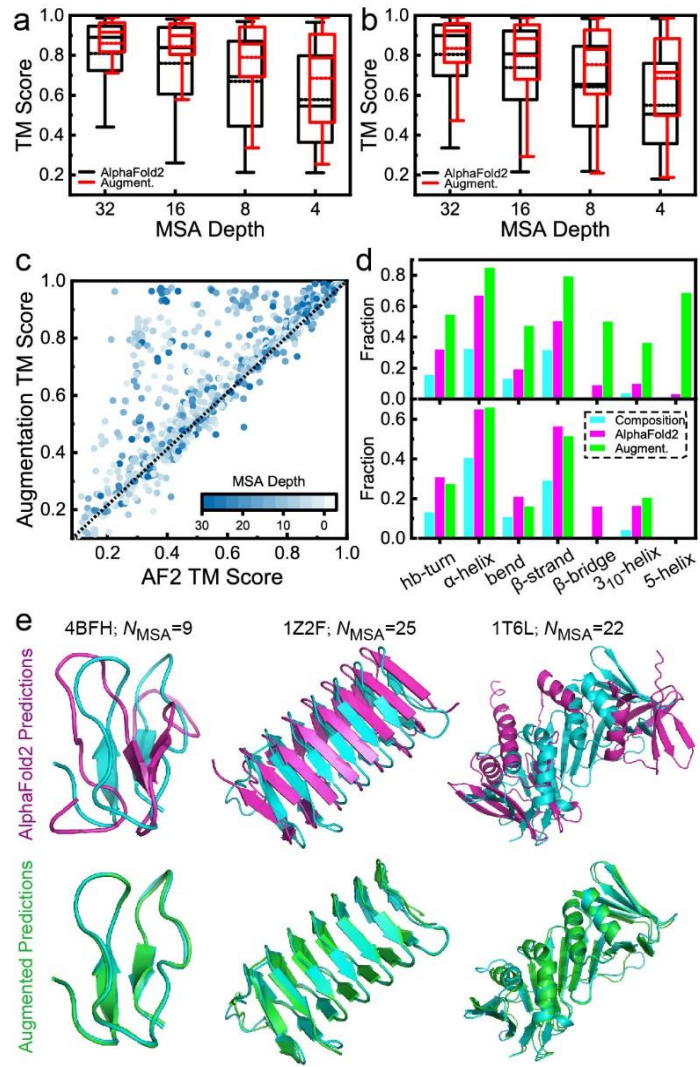

**Fig. 5 | Generative MSA augmentation improves few-shot structure prediction. a**, Performance of AF2 over CASP14 targets at varied MSA depths without (black boxes) and with (red boxes) MSA augmentation. **b**, Performance of AF2 over CAMEO targets at varied MSA depths without (black boxes) and with (red boxes) MSA augmentation. **c**, The quality of predicted structures with and without MSA augmentation over poor MSA targets. Only data points with |ΔTMScore|>0.01 are shown for better visualization. Each data point is colored according to available MSA depth during inference. **d**, Statistics of correctly predicted secondary structures in "improved set" (upper panel; see definition in the main text) and "underperformed set" (lower panel; see definition in the main text). Green and magenta bars represent the fraction of correct predictions for AF2 with and without MSA augmentation, respectively; The composition of secondary structures of the entire set is shown in cyan. **e**, Comparison of structures predicted for 4BFH, 1Z2F and 1T6L (from left to right) without MSA augmentation (magenta), with MSA augmentation (green), and the ground truth (cyan).

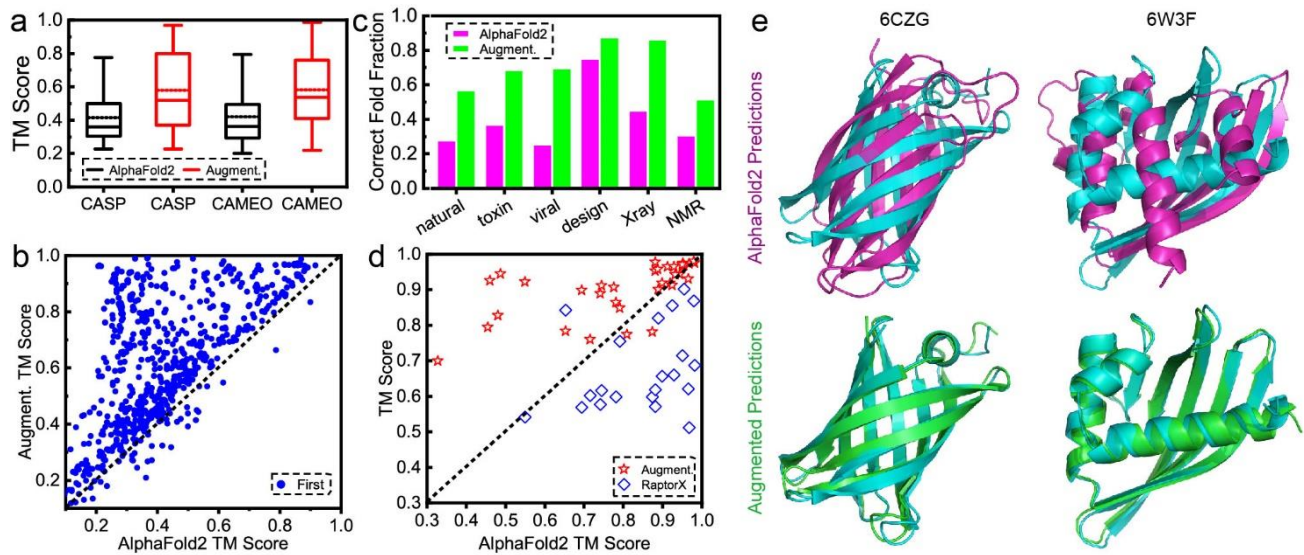

**Fig. 6 | EvoGen enables efficient single-sequence structure prediction. a**, Performance of single-sequence AF2 over CASP14 and CAMEO targets at varied MSA depths without (black boxes) and with (red boxes) MSA augmentation. **b**, The quality of single-sequence AF2 predictions with and without MSA augmentation over poor MSA targets. Only data points with $|\Delta TMScore|>0.05$ are shown for better visualization. **c**, Fraction of correctly folded structures (TMScore>0.5) via single-sequence AF2 for various types of proteins in poor MSA test set with (green) or without (magenta) MSA augmentation. **d**, The quality of single-sequence AF2 predictions with (red stars) and without MSA augmentation over de novo designed targets. Performance of RaptorX on applicable targets is also shown (blue squares) for comparison. **e**, Comparison of structures predicted for designed proteins 6CZG (left) and 6W3F (right) without MSA augmentation (magenta), with MSA augmentation (green), and the ground truth (cyan).

# Supplementary Information

## Unsupervisedly Prompting AlphaFold2 for Few-Shot Learning of Accurate Folding Landscape and Protein Structure Prediction

Jun Zhang[1,†], Sirui Liu[1], Mengyun Chen[2], Haotian Chu[2], Min Wang[2], Zidong Wang[2], Jialiang Yu[2], Ningxi Ni[2], Fan Yu[2], Dechin Chen[3], Yi Isaac Yang[3], Boxin Xue[4], Lijiang Yang[4], Yuan Liu[5] and Yi Qin Gao[1,3,4,6,†]

**Affiliations:**

[1] Changping Laboratory, Beijing 102200, China.

[2] Huawei Hangzhou Research Institute, Huawei Technologies Co. Ltd., Hangzhou 310051, China.

[3] Institute of Systems and Physical Biology, Shenzhen Bay Laboratory, Shenzhen 518055, China.

[4] Beijing National Laboratory for Molecular Sciences, College of Chemistry and Molecular Engineering, Peking University, Beijing 100871, China.

[5] Department of Chemical Biology, College of Chemistry and Molecular Engineering, Peking University, Beijing 100871, China.

[6] Biomedical Pioneering Innovation Center, Peking University, Beijing 100871, China.

† To whom correspondence should be addressed. Email: jzhang@cpl.ac.cn or gaoyq@pku.edu.cn

**Table of contents**

**Datasets**

1. Test sets

We prepared four independent test sets, i.e., CASP14, CAMEO, poor MSA and *de novo*, to benchmark performance of EvoGen. CASP14 test set contains 84 domain-divided single-chain targets in the official CASP14 name list with sequence length less or equal than 512. CAMEO test set contains all single-chain targets for CAMEO dating from 2021-08-21 to 2022-02-12. For ease of inference, we filtered out sequences longer than 512, resulting in 292 targets in total. Poor MSA dataset consists of single protein chains with known PDB structures but with less than 30 available MSA's. It is created by filtering all PDB entries in PSP Database (PSPD)[1] with a date truncation at 2020-05-14. Since AF2 is trained for single chain PSP, we further filtered this dataset to exclude any chains forming protein-protein interactions in heteromers. We also removed any sequences (with labeled structures) which are shorter than 15 amino acids. The resulting dataset contains 1074 targets, among which 382 targets do not have any MSA and are excluded for few-shot MSA augmentation experiments (Section III in the main text). Besides, we reused the list of de novo targets for RaptorX[2] which contains 35 artificial designed proteins using the Rosetta energy function. Twenty-one targets in this set were benchmarked by RaptorX, and we plotted the results of RaptorX on these applicable targets in Fig. 6d.

2. MSA trimming

Because all experiments in this paper were designed for low-data regime, we performed *MSA trimming* for CASP14 and CAMEO targets whenever MSA is abundant. Given a maximum MSA depth $N_{\max}$, the MSA trimming follows the same procedure as adopted for PSPD-Lite[1]. Specifically, for each target sequence whose MSA depth exceeding $N_{\max}$, we first filtered its MSA according to three primary rules: i) all MSA's with coverage less than 50% are removed; ii) all MSA's with >90% identity to target are removed; iii) all MSA's with <20% identity to target are removed. If MSA depth of the target still exceeds $N_{\max}$ after filtering, we further selected representative MSA's via a heuristic strategy as follows: We initialized an MSA pool using the target sequence alone, then added to this pool a new MSA given that this candidate is of no more than 90% identity to all MSA's already in the pool, and that this candidate is closest to the target in terms of the Hamming's distance. This iterative selection stops when no more candidates can be accepted or the MSA pool is full (up to $N_{\max}$). MSA trimming with $N_{\max}$=128 was performed for CASP14 and CAMEO test sets.

3. Training sets

We curated two training sets for EvoGen. The "labeled set" contains both sequences and structural labels, while the "unlabeled set" is composed merely of sequences (and MSA) without structural labels. The labeled set consists of 447K filtered PDB structures extracted from PSPD-Lite with a date truncation before 2020-05-14. CASP14 and CAMEO test sets are naturally excluded from the training set. The unlabeled set further expands the labeled set by adding 648K filtered non-redundant sequences in UniRef50[3] extracted from PSPD-Lite[1], and only the sequence information (i.e. MSA) is preserved whereas the structure labels are deprecated. For both training sets, MSA trimming is performed with a $N_{\max} = 256$ following the strategy described above. Additional filtering was performed after trimming: i) All entries with MSA depth smaller than 128 are removed (the poor MSA test set is thus excluded from the training set); ii) Any sequences in the de novo test set are manually removed; iii) Sequences or structural labels with length shorter than 20 amino acids are also removed. EvoGen was trained on the unlabeled set for MSA calibration, and fine-tuned on the labeled set for MSA augmentation.

**Training Settings**

1. Training objective of EvoGen

As elaborated in the main text, we aim to optimize the deep neural network model in order to maximize the conditional log-likelihood in Eq. (S1),

$$LL = \mathbb{E}_{m \in \mathcal{D}, \mathbf{S}_m \in \{\mathbf{S}_m\}_{\text{target}}} \log p_\theta \left( \mathbf{S}_m \mid \{\mathbf{S}_m\}_{\text{context}} \right) \quad (S1)$$

where we divide a full set of MSA $m$ into two subsets: $\{\mathbf{S}_m\}_{\text{context}}$ serves as conditional information, while $\{\mathbf{S}_m\}_{\text{target}}$ is used as training targets. This likelihood is intractable, however, we can derive an evidence lower bound (ELBO) for it by means of variational inference. Simply speaking, log-likelihood in Eq. (S1) can be re-formulated as Eq. (S2),

$$\log p \left( \mathbf{S}_m \mid \{\mathbf{S}_m\}_{\text{context}}, \theta \right) = \log \int p \left( \mathbf{S}_m, \mathbf{z} \mid \{\mathbf{S}_m\}_{\text{context}}, \theta \right) d\mathbf{z} \quad (S2)$$

where $\mathbf{z}$ is a latent variable generated by a (potentially data-dependent) prior, and it has a lower bound according to Jensen's equality where we denote $\{\mathbf{S}_m\}_{\text{context}}$ as $\{\mathbf{S}_m\}$ for short,

$$\log p\left(\mathbf{S}_m \mid \{\mathbf{S}_m\}, \theta\right) \geq \mathcal{L}(\theta, \phi)$$
$$\mathcal{L}(\theta, \phi) = \mathbb{E}_{q_\phi} \left[ \log p_\theta \left( \mathbf{S}_m, \mathbf{z} \mid \{\mathbf{S}_m\} \right) - \log q_\phi \left( \mathbf{z} \mid \mathbf{S}_m, \{\mathbf{S}_m\} \right) \right] \quad (S3)$$
$$= \mathbb{E}_{q_\phi} \left[ \log p_\theta \left( \mathbf{S}_m \mid \mathbf{z}, \{\mathbf{S}_m\} \right) - D_{\text{KL}} \left( q_\phi \left( \mathbf{z} \mid \mathbf{S}_m, \{\mathbf{S}_m\} \right) \| p_\theta \left( \mathbf{z} \mid \{\mathbf{S}_m\} \right) \right) \right]$$

Eq. (S3) consists of two models: a generative or decoder model $p_\theta$ performing reconstruction according to the context and latent variable $\mathbf{z}$, whereas an inference or encoder model $q_\phi$ performing variational inference for the posterior. The tightness of ELBO is controlled by the variational inference model $q_\phi$ which aims to minimize the Kullback-Leibler (KL) divergence to the true posterior $p(\mathbf{z} \mid \mathbf{S}_m, \{\mathbf{S}_m\})$. Therefore, it is natural to approximate both models with a deep neural network which is known for its expressivity as in Variational Auto-Encoders (VAE)[4]. During training, we optimized model parameters in order to maximize the ELBO in Eq. (S3).

In vanilla VAE, the prior for latent variable is usually a simple distribution like the standard normal. However in EvoGen, since we are dealing with contexts and sequences of varied lengths, we choose to learn a data-dependent prior for the latent variables.

Particularly, like denoising diffusion models[5], the dimension of latent variables is consistent with the length of target sequence, hence, making the model transferable to sequences of varied length. Besides, we introduced multi-scale priors which take the form of autoregressive Gaussians[6,7] to make ELBO tighter. Compared to a single Gaussian, autoregressive Gaussians can better approximate any complex distribution, meanwhile allow fast and straightforward sampling which is crucial to the selection of priors.

Note that Eq. (S3) consists of two terms, one for reconstruction loss as in an autoencoder, the other for KL divergence which can be considered as a regularizer. To stabilize training and avoid posterior collapse[8], we adopted a warm-up schedule as in NVAE[7] to gradually tune-up the strength of the KL divergence term. Besides, since both terms depend on the length of input sequence, we balanced the mini-batch gradient according to the sequence length as well. We scaled the loss of each MSA with a weight factor proportional to the square root of target length as recommended by AlphaFold2[9]. We trained the model using a batch size of 128 MSAs, each MSA was cropped to a maximum length of 256 and maximum depth of 128. We adopted ADAM optimizer[10] (with default beta, epsilon=1e-6) and clipped the gradient by norm bounded by 0.1. The learning rate was warmed up from 0 to 5e-4 during the first 3K steps, then decayed according to a cosine learning rate schedule to 1e-5 during 100K steps. In total 150K training iterations (or gradient steps) were executed for unsupervised pre-training using the unlabeled training dataset (see Datasets in SI) which aims to maximize Eq. (S3), and the resulting model is used for MSA calibration throughout the paper.

Another 50K training iterations were performed using the labeled training dataset (see Datasets in Supplementary Information) under the guidance of AF2 which aims to minimize the combined loss in Eq. (S4),

$$\mathcal{L}_{\text{finetune}} = 0.5\mathcal{L}_{\text{FAPE}} + 0.5\mathcal{L}_{\text{torsion}} + 0.01\mathcal{L}_{\text{viol}} + 0.01\mathcal{L}_{\text{conf}} - 0.1\mathcal{L}_{\text{EvoGen}} \quad (S4)$$

where $\mathcal{L}_{\text{FAPE}}$ stands for clamped frame-aligned point errors (FAPE) of both backbone and sidechains, $\mathcal{L}_{\text{torsion}}$ for the loss of sidechain torsional angles, $\mathcal{L}_{\text{viol}}$ for violation losses, $\mathcal{L}_{\text{conf}}$ for confidence loss, and $\mathcal{L}_{\text{EvoGen}}$ corresponds to Eq. (S3). All loss terms in Eq. (S4) except $\mathcal{L}_{\text{EvoGen}}$ take the same form as AF2[9]. Note that we also relaxed the parameters of Evoformer module and the confidence head of AF2 during fine-tuning since we observed that virtual MSA generated by EvoGen may cause AF2 to overestimate the quality of predictions. The fine-tuned model was adopted for MSA augmentation throughout the paper.

We performed training over 128 Ascend-910 NPU cards using MindSpore[11] and adopted hybrid float precisions during training to reach the optimal performance.

## 2. Differentiate through AF2

For MSA augmentation, we trained EvoGen with respect to relevant structural losses feedbacked by AF2 according to Eq. (S4). However, to compute the supervised losses, we need to transform the softmax-valued output (see "Model Details" in SI) of EvoGen to be one-hot-coded MSA features then passed to AF2. Simply using ArgMax transform would stop the gradient and forbid backpropagation through EvoGen.

Therefore, we applied Gumbel-Softmax trick[12] to generate nearly one-hot samples according to the softmax logits, and adopted straight-through estimator[13,14] to allow backpropagation of EvoGen in joint with AF2. Let $\mathbf{S}_{gs}$ denote Gumbel-Softmax samples which are differentiable with respect to EvoGen parameters, $\mathbf{S}_{hard}$ denote one-hot MSA features after ArgMax transform of $\mathbf{S}_{gs}$, and $f(\mathbf{S}_{gs})$ is an arbitrary function of EvoGen output, the straight-through estimator reads like Eq. (S5),

$$f(\mathbf{S}) = f(\mathbf{S}_{gs}) + \text{StopGrad}\left[f(\mathbf{S}_{hard}) - f(\mathbf{S}_{gs})\right] \quad (S5)$$

where "StopGrad" stands for stop-gradient operation. During forward inference, Eq. (S5) computes the function value using the one-hot coded $\mathbf{S}_{hard}$, while during backpropagation, the gradient with respect to the Gumbel-Softmax samples are computed.

## Inference Settings

<u>1. Inference settings of AF2</u>

We conducted all the experiments without templates. AF2 model-3 released by DeepMind was chosen for inference and training in all experiments unless specified otherwise. AF2 model-3 slightly outperformed the other two template-free models (model-4 and model-5) on our benchmark dataset, and it is also recommended as default model by batch-mode ColabFold[15]. After MSA trimming, MSA subsampling is no longer performed during inference, except when we deliberately sub-sampled MSA for purposes (see Section I in Experiments & Results). We also turned off any other settings which could cause non-deterministic effects (e.g., BERT) during AF2 inference. Unless stated otherwise, AF2 inference was executed exclusively using a recommended number of three recycles.

<u>2. Inference settings of EvoGen</u>

One special hyper-parameter during EvoGen inference (for both MSA calibration and augmentation) is the context MSA ratio, which determines how much fraction of available MSA is used as contexts during inference. Let $r_{\text{ctx}}$ denotes context MSA ratio range between 0 and 1, and $N_{\text{MSA}}$ denote the available MSA number (possibly after MSA trimming) provided to EvoGen, then the number of context MSA is the integer part of $r_{\text{ctx}} N_{\text{MSA}}$. Note that regardless of $r_{\text{ctx}}$, the first sequence in MSA, i.e., the query sequence itself, is always included in the context.

By setting a large $r_{\text{ctx}}$ (close to unity), the calibrated or generated MSA tend be more consistent. In contrast, a small $r_{\text{ctx}}$ means more randomness in sub-sampled contexts, and usually leads to more noisy output. This hyper-parameter can help us strike balance between exploration (with smaller $r_{\text{ctx}}$) and exploitation (with larger $r_{\text{ctx}}$). In our experiments, we chose three values for $r_{\text{ctx}} \in \{0.5, 0.7, 0.9\}$ for each all tasks where multiple MSA sequences are available unless specified otherwise.

For MSA augmentation, there is an additional hyper-parameter $N_{\text{aug}}$ controlling the augmented MSA depth. For few-shot learning, we set $N_{\text{aug}} = 128$ in order to make a fair comparison to vanilla AF2 with trimmed MSA depth of 128. For single-sequence prediction, or zero-shot learning, we ran inference using three different values $N_{\text{aug}} \in \{16, 32, 64\}$, to test the impact of this hyper-parameter, and did not observe significant change of performance as long as $N_{\text{aug}} \geq 32$.

After finetuned under the guide of AF2, we found that directly fed Softmax output of EvoGen without any hardening transform to the downstream AF2 model yields slightly better performance for MSA augmentation. This might benefit from the "dark knowledge" in the Softmax output which turns a token of amino acid (one-hot code) at a position into a distribution of all possible amino acids at this position, hence, helps smooth the folding landscape of AF2.

Given a specific choice of $r_{\text{ctx}}$ (and) or $N_{\text{aug}}$, we ran five independent inferences using different Gaussian random noises for MSA calibration and few-shot MSA augmentation experiments. In zero-shot MSA augmentation experiments, we reduced the number of random trials to two. Among all executed trials, we ranked all predictions according to the confidence score (i.e., residue-averaged pIDDT) yielded by AF2, and reported the top-1 prediction as the "first prediction" in all experiments. We also reported the *de facto* "best prediction" with the ground truth label as reference.

We recorded all the "first" and "best" predictions in our experiments, which can be checked via the open-source link. We also kept records of the output of EvoGen (i.e., calibrated MSA features) which could be used to reproduce the reported structures using third-party implementation of AF2 like ColabFold[15].

3. Probing alternative conformations

When MSA is sufficiently deep, direct implementing AF2 inference will lead to limited variations in predicted structures as proved in this paper and related work[16]. Consequently, implementing the generative inference workflow presented in this paper to probe alternative may find wide applications in protein science beyond few-shot learning scenarios.

We summarized a brief protocol of how to increase the diversity of AF2 prediction with the help of EvoGen. First, select a reasonable $N_{\text{max}}$ and perform MSA trimming accordingly. Secondly, randomly sub-sample $N_{\text{sub}}$ from the trimmed MSA pool and feed them to EvoGen. We remark here that previous research[16] also suggested implementing AF2 with a shallow MSA in order to get diverse structure predictions. Thirdly, choose a context MSA ratio $r_{\text{ctx}}$ and perform MSA calibration accordingly with one or more random seeds. Finally, pass the reconstructed MSA features to AF2 and perform structure predictions, and cluster the confident predictions with proper similarity metrics like TMScore[17]. According to our experiments, we recommend $N_{\text{max}} = 512$ or $1024$, $N_{\text{sub}} \in \{16,32,64\}$ and $r_{\text{ctx}} \in \{0.25,0.5,0.75\}$ in practice for efficient probing of alternative protein conformations.

## Model Details

1. Input and output of EvoGen

The input to EvoGen is a set of MSA sequences $m \equiv \{\mathbf{S}_m^i\}_{i=1,\ldots,N_m}$. The first sequence is always the query sequence ($S_m^1 \equiv Q_m$), the query sequence does not contain gaps or deletions. While the other sequences are aligned to the query, they may contain gaps or deletions due to alignment.

Each sequence $\mathbf{S}_m^i \in m$ is featurized by the type of amino acid and the number of deletions at each position along the sequence. The amino acid is categorized into a vocabulary of 22 tokens, including 20 for common amino acids, 1 for rare amino acids and 1 for gap token. The deletion number of each position in a sequence is transformed via arctan function as in AF2.

The output of EvoGen should correspond to the input in order to perform reconstruction. The amino acid type at each position along the sequence is predicted by a softmax function with 22 logits corresponding to vocabulary tokens. The arctan deletion number is first discretized into 6 bins ranging from 0.2 to 0.95, and a softmax function with 6 logits predicts the discretized values.

2. Hyperformer

Hyperformer inherits the overall architecture of Evoformer[9] but exhibits several key differences (Fig. 1c). First of all, the original biased attention is replaced with hyper-attention inspired by Molecular CT[18], and the attention coefficient between a $d$-dimensional Query vector $\mathbf{q}_i$ and Key vector $\mathbf{k}_j$ is computed as

$$\text{Att}(i,j) = \text{softmax}\left(\frac{\mathbf{q}_i^T \mathbf{W}_{ij} \mathbf{k}_j}{\sqrt{d}} + b_{ij}\right) \qquad (S6)$$

where $\mathbf{W}_{ij}$ and $b_{ij}$ are learnable parameters or activations of neural networks which are both functions of the relative positions (or pair activations) between $i$-th and $j$-th tokens. Similar to hyper-networks[19], $\mathbf{W}_{ij}$ and $b_{ij}$ here represent learnable affine transform of the space basis and the offset of the resulting inner product, respectively. Vanilla attention (or biased attention) is a special case of hyper-attention in Eq. (S6) given an identity $\mathbf{W}_{ij}$ and zero (or non-zero) $b_{ij}$. In EvoGen we adopted rotary positional embedding (RoPE)[20] as $\mathbf{W}_{ij}$, so that $\mathbf{W}_{ij}$ can be decomposed into product of two position-dependent vectors and merged with the linear transform of Query and Key vectors. The second difference lies in the embedding of relative positions. We adopted an approach similar to T5 model[21] and grouped $|i - j|$ into discretized buckets according to log-scales[22]. This way of relative positional embedding not only expands the horizon of sequence models without inducing extra memory cost, but also equips the model with a hyperbolic view of distances as inductive biases. Thirdly, we added a new Query Conditioning Module into EvoGen encoder (Fig. 1c), which is a neural network that mixes context activations with query activations in order to help the model learn relative differences between MSA and the target sequence. Lastly, similar to AlphaFold-Multimer[23], we changed the order of the "outer product mean" operation to the beginning of each Hyperformer block (Fig. 1c), allowing the single update and pair update to be executed in parallel and separately.

3. Latent module

Latent module is designed to summarize the statistics of MSA features. Latent modules in encoder (matching network 1 in Fig. 1d) are responsible for summarizing target sequences into deviations with respect to the data-dependent priors, then the posteriors are calculated as the addition of priors and the corresponding deviations. Such posterior formula reflects the principle of "relativity" in the model design and stabilizes training, which was first observed in NAVE[7]. On the other hand, latent modules in decoder (matching network 2 in Fig. 1d) are responsible for estimating the priors according to context sequences.

Given the overall symmetry between encoder and decoder (Fig. 1a) and the principle of "relativity", we employed twin (or Siamese) matching networks to learn the relevant statistics (Fig. 1d). Another neural network called Sampling Module, draws random samples according to learned posteriors with Gaussian noises via re-parametrization trick[4]. During generation, only context sequences are provided to EvoGen and the model predicts the priors, according to which we can sample new sequences.

4. Model hyperparameters

EvoGen is composed of a pair of encoder and decoder with relative symmetry similar to U-Net[24], each consisting of 12 Hyperformer blocks. Similar to AF2, in Hyperformer, we set the dimension of sequence representation to be 256 and the dimension of pair representation to be 128 (Fig. 1b in the main text). Therefore, the scaling parameter for hyper-attention $d = 256$ in Eq. (S6). According to the principle of "hierarchy", we adopted 3 Latent Module blocks between the encoder and decoder (Fig. 1a), with increasing latent dimensions (64, 128, 256, respectively) during encoding (or equivalently, decreasing dimensions during decoding).